%% file: neurips_2020.tex
\documentclass{article}

\PassOptionsToPackage{numbers}{natbib}

     \usepackage[preprint]{neurips_2020}

\usepackage[utf8]{inputenc} 
\usepackage[T1]{fontenc}    
\usepackage{hyperref}       
\usepackage{url}            
\usepackage{booktabs}       
\usepackage{amsfonts}       
\usepackage{nicefrac}       
\usepackage{microtype}      

\usepackage{graphicx}
\usepackage{wrapfig}
\usepackage{subfig}
\usepackage{multirow}
\usepackage{tabularx}
\usepackage{siunitx}
\usepackage{lipsum}

\usepackage[table]{xcolor}
\definecolor{nicergreen}{rgb}{0.13, 0.54, 0.13}
\definecolor{nicered}{rgb}{0.83, 0.16, 0.16}
\newcommand\showdiff[1]{\textbf{\textcolor{nicergreen}{#1}}}
\newcommand\showdiffn[1]{\textbf{\textcolor{nicered}{#1}}}

\newcolumntype{C}{>{\centering\arraybackslash}X}

\usepackage[super]{nth}

\title{Are we done with ImageNet?}

\author{%
  \hspace{-15pt}
  Lucas Beyer$^1$\thanks{All authors contributed equally, project led by first author.
  \newline
  \hspace*{1.8em}Correspondence: \{lbeyer, henaff, akolesnikov, xzhai, avdnoord\}@google.com}~~
  Olivier J. Hénaff$\,^2$\footnotemark[1]~~
  Alexander Kolesnikov$^1$\footnotemark[1]~~
  Xiaohua Zhai$^1$\footnotemark[1]~~
  Aäron van den Oord$^2$\footnotemark[1] \\
  $^1$Google Brain (Zürich, CH) and $^2$DeepMind (London, UK)
\setcounter{footnote}{0}
}

\begin{document}

\maketitle

\setcounter{footnote}{2}

\begin{abstract}
Yes, and no. We ask whether recent progress on the ImageNet classification benchmark continues to represent meaningful generalization, or whether the community has started to overfit to the idiosyncrasies of its labeling procedure. We therefore develop a significantly more robust procedure for collecting human annotations of the ImageNet validation set. Using these new labels, we reassess the accuracy of recently proposed ImageNet classifiers, and find their gains to be substantially smaller than those reported on the original labels. Furthermore, we find the original ImageNet labels to no longer be the best predictors of this independently-collected set, indicating that their usefulness in evaluating vision models may be nearing an end. Nevertheless, we find our annotation procedure to have largely remedied the errors in the original labels, reinforcing ImageNet as a powerful benchmark for future research in visual recognition\footnote{The new labels and rater answers are available at https://github.com/google-research/reassessed-imagenet}.
\end{abstract}

\begin{wrapfigure}{r}{0.45\textwidth}
  \vspace{-6em}
  \begin{center}
    \includegraphics[width=0.45\textwidth]{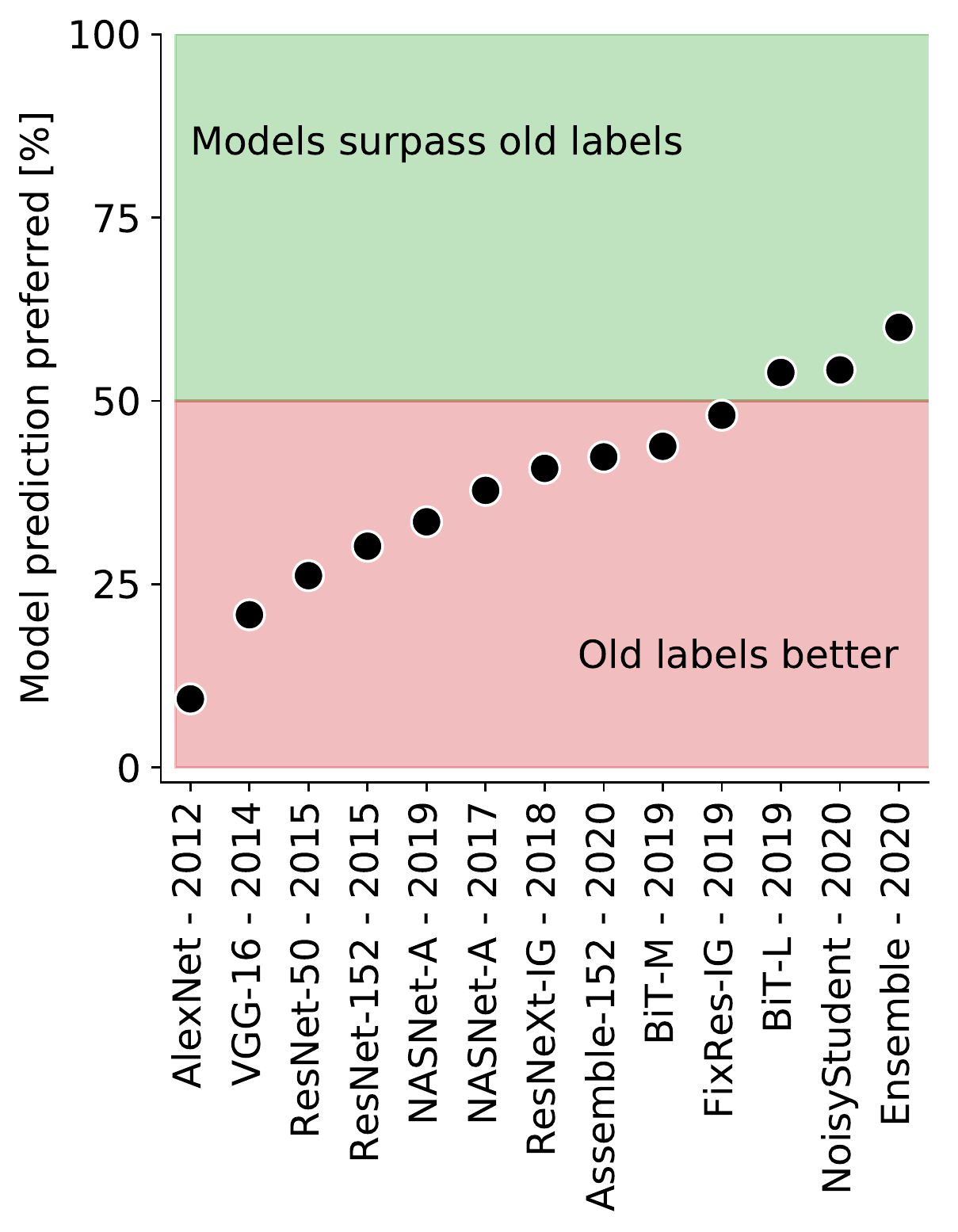}
  \end{center}
  \vspace{-1em}
  \caption{When presented with a model's prediction and the original ImageNet label, human annotators now prefer model predictions on average (Section~\ref{sec:reeval}). Nevertheless, there remains considerable progress to be made before fully capturing human preferences.} 
  \label{fig:teaser}
  \vspace{-2em}
\end{wrapfigure}

\section{Introduction}
\label{sec:intro}

For nearly a decade, the ILSVRC image classification benchmark~\cite{russakovsky2015imagenet}, ``ImageNet'', has been a central testbed of research in artificial perception.
In particular, its scale and difficulty have highlighted landmark achievements in machine learning, starting with the breakthrough AlexNet~\cite{krizhevsky2012imagenet}.
Importantly, success on ImageNet has often proven to be general: techniques that advance its state-of-the-art have usually been found to be successful in other tasks and domains. For example, progress on ImageNet due to architecture design~\cite{krizhevsky2012imagenet,simonyan2014very,he2016deep} or optimization~\cite{ioffe2015batch} has yielded corresponding gains on other tasks and modalities~\cite{vaswani2017attention, oord2016wavenet, silver2017mastering}. Similarly, successful representation learning techniques applied to ImageNet~\cite{oord2018representation, caron2018deep, zhuang2019local} have yielded corresponding gains elsewhere~\cite{han2019video, schneider2019wav2vec, alwassel2019self, zhuang2019unsupervised}.

As recent results continue to report systematic gains on ImageNet, we ask whether this progress continues to be as general as before. 
We therefore develop a method for collecting human annotations which leverages the predictions of a diverse set of image models, and use these Reassessed Labels (``ReaL'') to re-evaluate recent progress in image classification.
While early progress on ImageNet almost perfectly translates into corresponding gains in ReaL accuracy, we find this association to be significantly weaker for more recent models.
Surprisingly, we find the original ImageNet labels to no longer be the best predictors of our newly-collected annotations, with recent high-performing models systematically outperforming them (Figure~\ref{fig:teaser}).
Next, we analyse the discrepancies between ImageNet and ReaL accuracy, finding that some of the ``progress'' on the original metric is due to overfitting to the idiosyncrasies of their labeling pipeline.
Finally, we leverage these observations to propose two simple techniques which address the complexity of ImageNet scenes, leading to systematic gains in both ImageNet and ReaL accuracy.

\textbf{Related work.} Several studies have revisited common computer vision benchmarks~\cite{radenovic2018revisiting,barz2019we,weyand2020google,musgrave2020metric} and, regarding ImageNet~\cite{russakovsky2015imagenet} specifically, identified various sources of bias and noise~\cite{northcutt2019confident,hooker2019selective,stock2018convnets}.
However, none of these investigate the effect of ImageNet's shortcomings, in particular how they might affect model accuracies and conclusions being drawn from them.
The most related recent work~\cite{recht2019imagenet} attempts to replicate the ImageNet collection pipeline, which results in a 12\% drop in accuracy.
However, they eventually conclude that accuracy gains on the original validation set perfectly translate to their newly collected data.
Later, \cite{engstrom2020identifying} demonstrate that this discrepancy is explained by a statistical bias in the data collection process that was unintentionally introduced in the replication.
Regardless, we significantly differ from this line of work, as we do not collect new images, but rather identify shortcomings of the existing validation labels and collect a new label set which addresses these.
ObjectNet~\cite{barbu2019objectnet} is another work that collects a new validation set that can be used to evaluate ImageNet models, focusing on gathering challenging ``real-life'' images in diverse contexts.
It also significantly differs from our work, as ObjectNet images introduces a strong domain shift, which is orthogonal to our enquiry.
Concurrently with our work, \cite{tsipras2020imagenet} proposes an improved pipeline for collecting ImageNet labels, although their analysis and conclusions differ from ours.

\section{What is wrong with ImageNet labels?}
\label{sec:whats-wrong}

The ImageNet dataset is a landmark achievement in the evaluation of machine learning techniques.
In particular, the ImageNet labeling pipeline allows for human judgments in a 1000-way real-world image classification task, an intractable problem using a naive interface.
Nevertheless, this procedure has some limitations which we seek to identify. 
Many images in the ImageNet dataset contain a clear view on a single object of interest: for these, a single label is an appropriate description of their content.
However many other images contain multiple, similarly prominent objects, limiting the relevance of a single label.
Even for images containing a single object, biases in the collection procedure can lead to systematic inaccuracies.
Finally, some ImageNet classes are inherently ambiguous, drawing distinctions between essentially identical groups of images.
In this section we review these sources of label noise, which motivate the design of a new human annotation procedure. 

\begin{figure}[h]
  \centering
  \includegraphics[width=\textwidth]{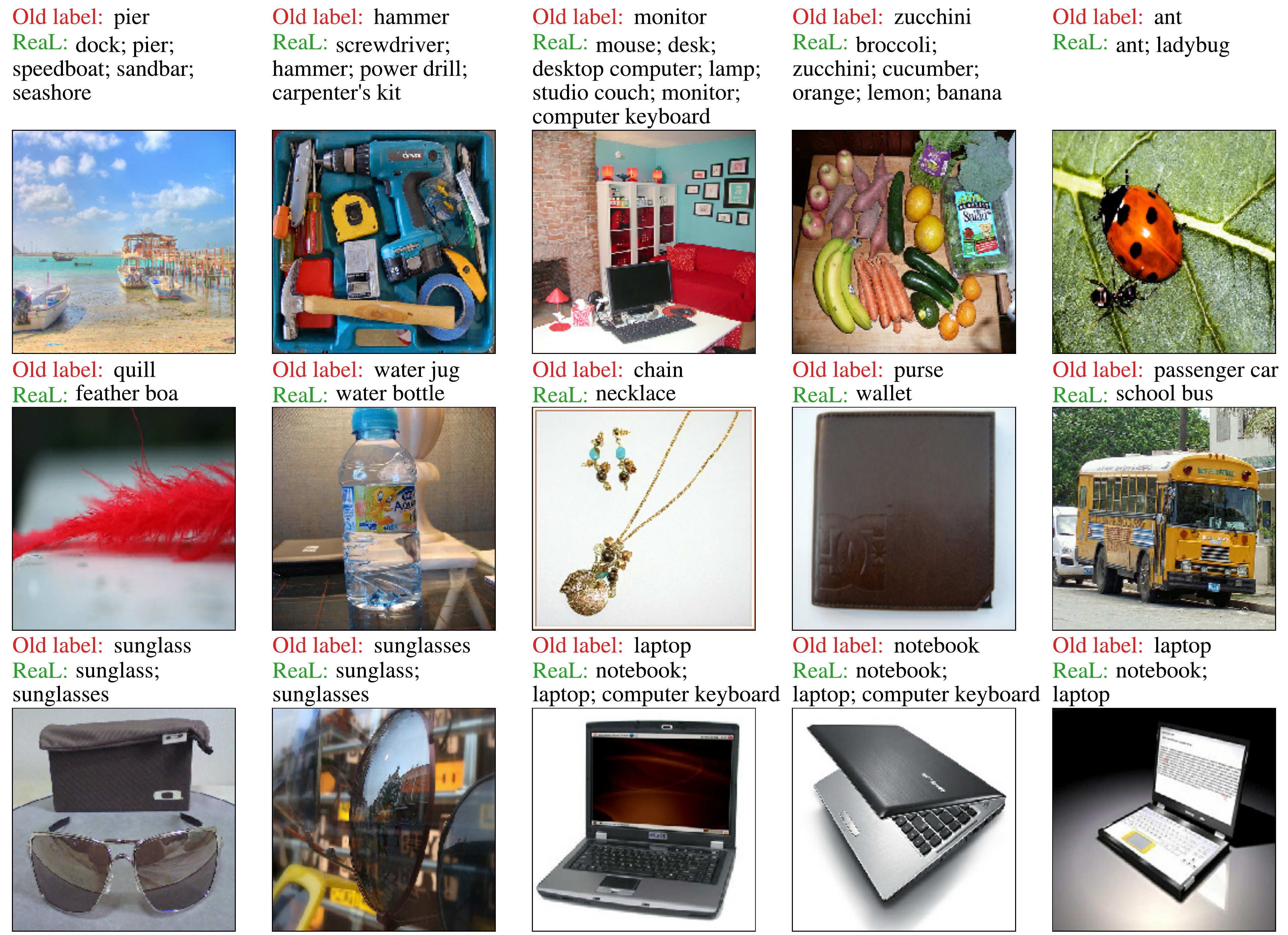}
  \caption{Example failures of the ImageNet labeling procedure. Red: original ImageNet label, green: proposed ReaL labels.
  \textbf{Top row}: ImageNet currently assigns a single label per image, yet these often contain several equally prominent objects.
  \textbf{Middle row}: Even when a single object is present, ImageNet labels present systematic inaccuracies due to their labeling procedure.
  \textbf{Bottom row}: ImageNet classes contain a few unresolvable distinctions.
}\label{fig:example-images}
  \vspace{-1em}
\end{figure}

\textbf{Single label per image.} Real-world images often contain multiple objects of interest.
Yet, ImageNet annotations are limited to assigning a single label to each image, which can lead to a gross under-representation of the content of an image (Figure \ref{fig:example-images}, top row).
In these cases, the ImageNet label is just one of many equally valid descriptions of the image, chosen in a way that reflects the idiosyncrasies of the labeling pipeline more than the content of the image.
As a result, using ImageNet validation accuracy as a metric can penalize an image classifier for producing a correct description that happens to not coincide with that chosen by the ImageNet label.
This motivates re-annotating the ImageNet validation set in a way that captures the diversity of image content in real-world scenes.

\textbf{Overly restrictive label proposals.} The ImageNet annotation pipeline consists of querying the internet for images of a given class, then asking human annotators whether that class is indeed present in the image.
While this procedure yields reasonable descriptions of the image, it can also lead to inaccuracies.
When considered in isolation, a particular label proposal can appear to be a plausible description of an image (Figure \ref{fig:example-images}, middle row).
Yet when considered together with other ImageNet classes, this description immediately appears less suitable (the ``quill'' is in fact a ``feather boa'', the ``passenger car'' a ``school bus'').
Based on this observation, we seek to design a labeling procedure which allows human annotators to consider (and contrast) a wide variety of potential labels, so as to select the most accurate description(s). 

\textbf{Arbitrary class distinctions.} ImageNet classes contain a handful of \emph{essentially} duplicate pairs, which draw a distinction between semantically and visually indistinguishable groups of images (Figure \ref{fig:example-images}, bottom row).
For example, the original ImageNet labels distinguish ``sunglasses'' from ``sunglass'', ``laptop'' from ``notebook'', and ``projectile, missile'' from ``missile''.
By allowing multiple annotations from simultaneously-presented label proposals, we seek to remove this ambiguity and arrive at a more meaningful metric of classification performance.

\section{Relabeling the ImageNet validation set}

Given the biases arising from assigning a single label in isolation, we design a labeling procedure which captures the diversity and multiplicity of content in the ImageNet dataset.
In particular, we seek a paradigm which allows human annotators to simultaneously evaluate a diverse set of candidate labels, while keeping the number of proposals sufficiently small to enable robust annotations.

\subsection{Collecting a comprehensive set of proposals}
\label{sec:model-selection}

We start by assembling a set of 19 models~\cite{simonyan2014very,szegedy2016rethinking,he2016deep,xie2017aggregated,mahajan2018exploring,kolesnikov2019big,lee2020compounding,zoph2018learning,cai2019once,zhai2019s4l,henaff2019data,chen2020improved,chen2020simple} which vary along a number of dimensions, including architecture, training objective and usage of external data (see appendix~\ref{sec:proposals}).
We use either a canonical public implementation, or predictions provided by the authors.
In order to generate label proposals from these models we adopt the following strategy: for any given image, we always include the original ImageNet (ILSVRC-2012) label and the Top-1 prediction of each model.
Moreover, given that many images may have more than one label, we allow each model to make additional proposals based on the logits and probabilities it assigns across the entire dataset (see appendix \ref{sec:proposals}).
We find this procedure for generating label proposals to result in near perfect label coverage, at the expense of having a very large number of proposed labels per image.

In order to reduce the number of proposals, we seek a small subset of models whose proposals retain a near perfect label coverage.
In order to quantify the precision and recall of a given subset, we had 256 images labeled by 5 computer vision experts, which we use as \emph{gold standard}.
We then perform an exhaustive search over model subsets to find a group which achieves the highest precision, while maintaining a recall above 97\%.
Based on this, we find a subset of 6 models which generates label proposals that have a recall of 97.1\% and a precision of 28.3\%, lowering the average number of proposed labels from 13 per image to 7.4 per image.
From this subset, we generate proposal labels for the entire validation set, using the same rule as described above.

\subsection{Human evaluation of proposed labels}

Having obtained a new set of candidate labels for the entire validation set, we start by assessing which images need to be evaluated by human raters.
In the event that all models agree with the original ImageNet label, we can safely retain the original without re-evaluation.
This reduces the number of images to be annotated from 50\,000 to 24\,889.
To avoid overwhelming raters with choices, we further split images with more than 8 label proposals into multiple labeling tasks according to the WordNet hierarchy.
This results in 37\,988 labeling tasks.

Each task is performed by 5 separate human annotators, using a crowdsourcing platform.
On a given trial a rater is presented with a single image and up to 8 candidate labels, and asked whether each label is present in the image (Figure~\ref{fig:precrec}, left).
For each label, the rater is instructed to respond {\tt yes} if they are 95\% sure that the label is indeed present in the image, {\tt no} if they are 95\% sure that the label is not present in the image, and {\tt maybe} otherwise.

\subsection{From human ratings to labels and a metric}

\begin{figure}%
    \centering
    \raisebox{-0.47\height}{\includegraphics[width=0.55\textwidth]{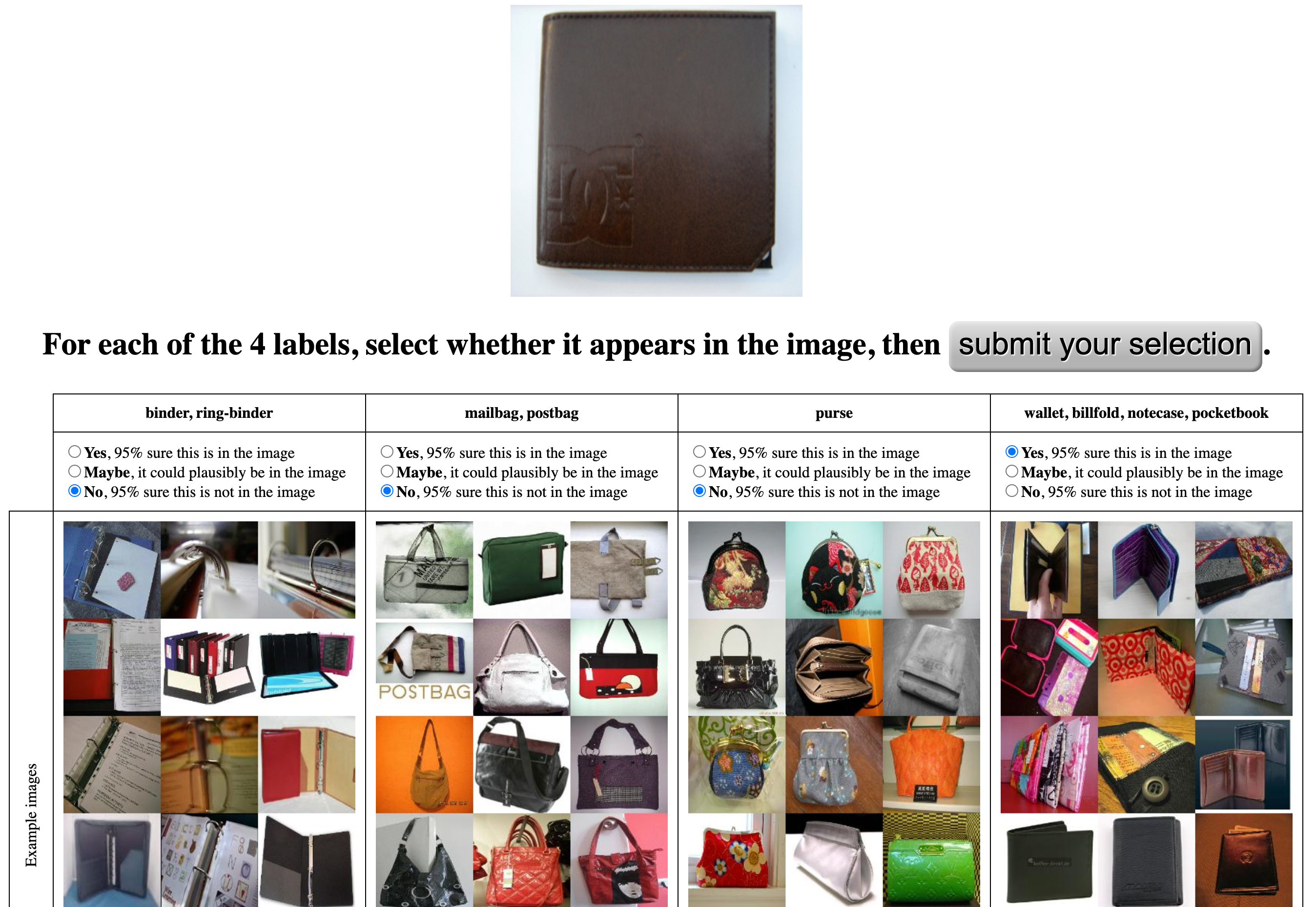}}%
    \hspace*{.1in}
    \raisebox{-0.5\height}{\includegraphics[width=0.45\textwidth]{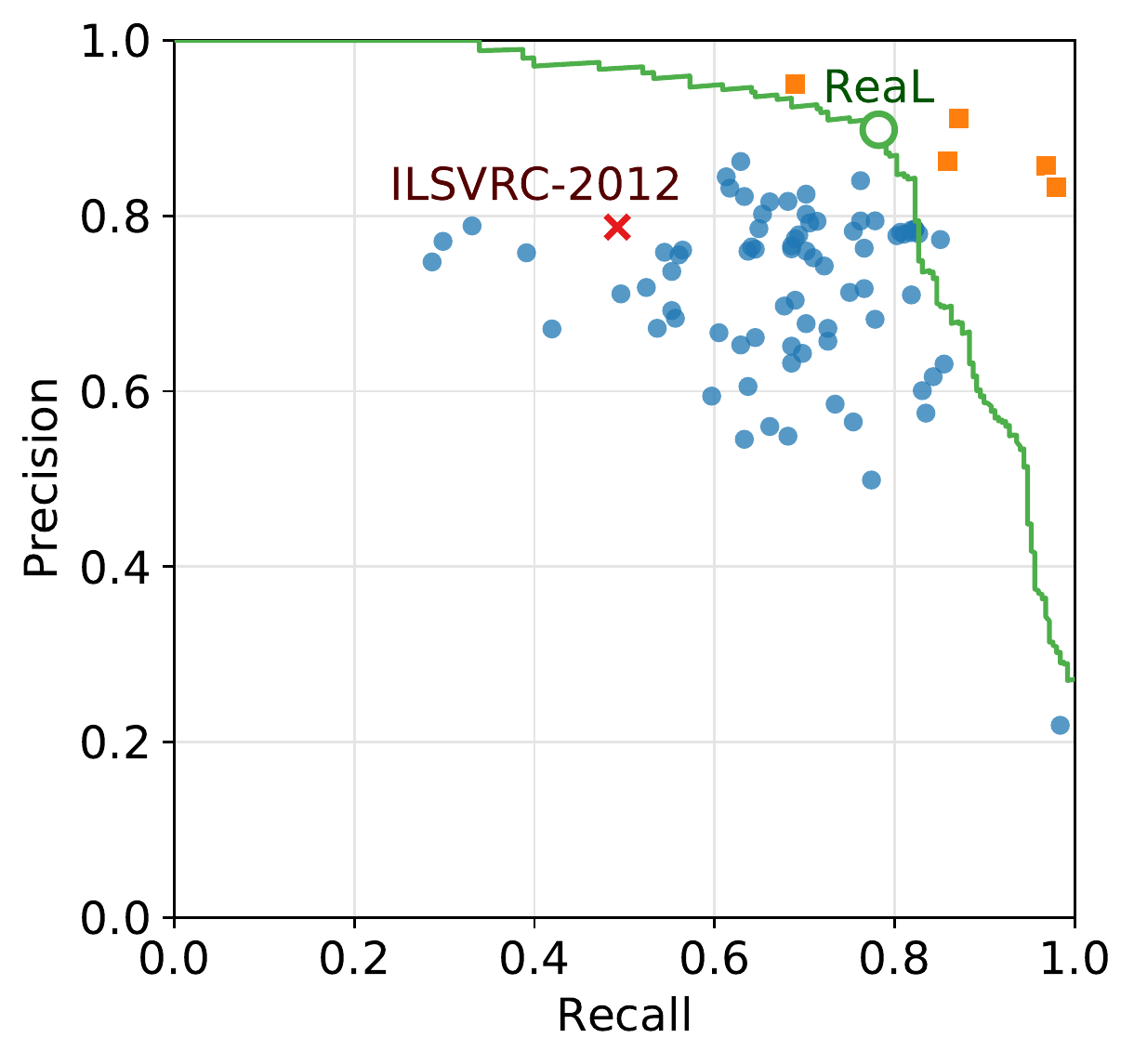}}%
    \vspace{-1em}
    \caption{Human assessment of proposed labels. \textbf{Left}: the interface for simultaneous evaluation of multiple label proposals. \textbf{Right}: precision and recall of individual raters (blue dots), expert annotators (orange boxes), and the model fit to rater data (green curve), evaluated on a majority-vote from expert annotators. The labels we use for subsequent analyses are indicated by a green circle, the original ImageNet labels with a red cross.}%
    \vspace{-1em}
    \label{fig:precrec}%
\end{figure}

The next step is to combine the 5 human assessments of every proposed label into a single binary decision of whether to retain or discard it.
In order to account for raters' varying characteristics, we use the classic method by Dawid and Skene~\cite{dawid1979maximum} which infers individual rater's error rates via maximum-likelihood.
We note that animal classes often carry more uncertainty and require expert opinion, which the original ImageNet labels indirectly incorporate via expert websites.
We therefore incorporate the ImageNet label as a virtual \nth{6} rater for images originally labeled as animals.
The precision-recall curve resulting from the whole process is shown in Figure~\ref{fig:precrec} (right).
We also tried simple majority voting, which performs well but worse: 90.8\% precision and 71.8\% recall.
The operating point we choose, marked by a circle on the curve, results in 57\,553 labels for 46\,837 images.
We discard the 3163 images that are assigned no label.

Equipped with these new validation set labels we propose a new metric: the ReaL accuracy (from Reassessed Labels), which addresses some of the shortcomings of the original ImageNet accuracy identified in Section~\ref{sec:whats-wrong}.
In particular, we wish to no longer penalise models for predicting one of multiple plausible labels of an image.
We therefore measure the precision of the model's top-1 prediction, which is deemed correct if it is included in the set of labels, and incorrect otherwise.

\section{Re-evaluating the state of the art}
\label{sec:reeval}

\begin{wrapfigure}{r}{0.55\textwidth}
  \vspace{-2em}
  \begin{center}
    \includegraphics[width=0.54\textwidth]{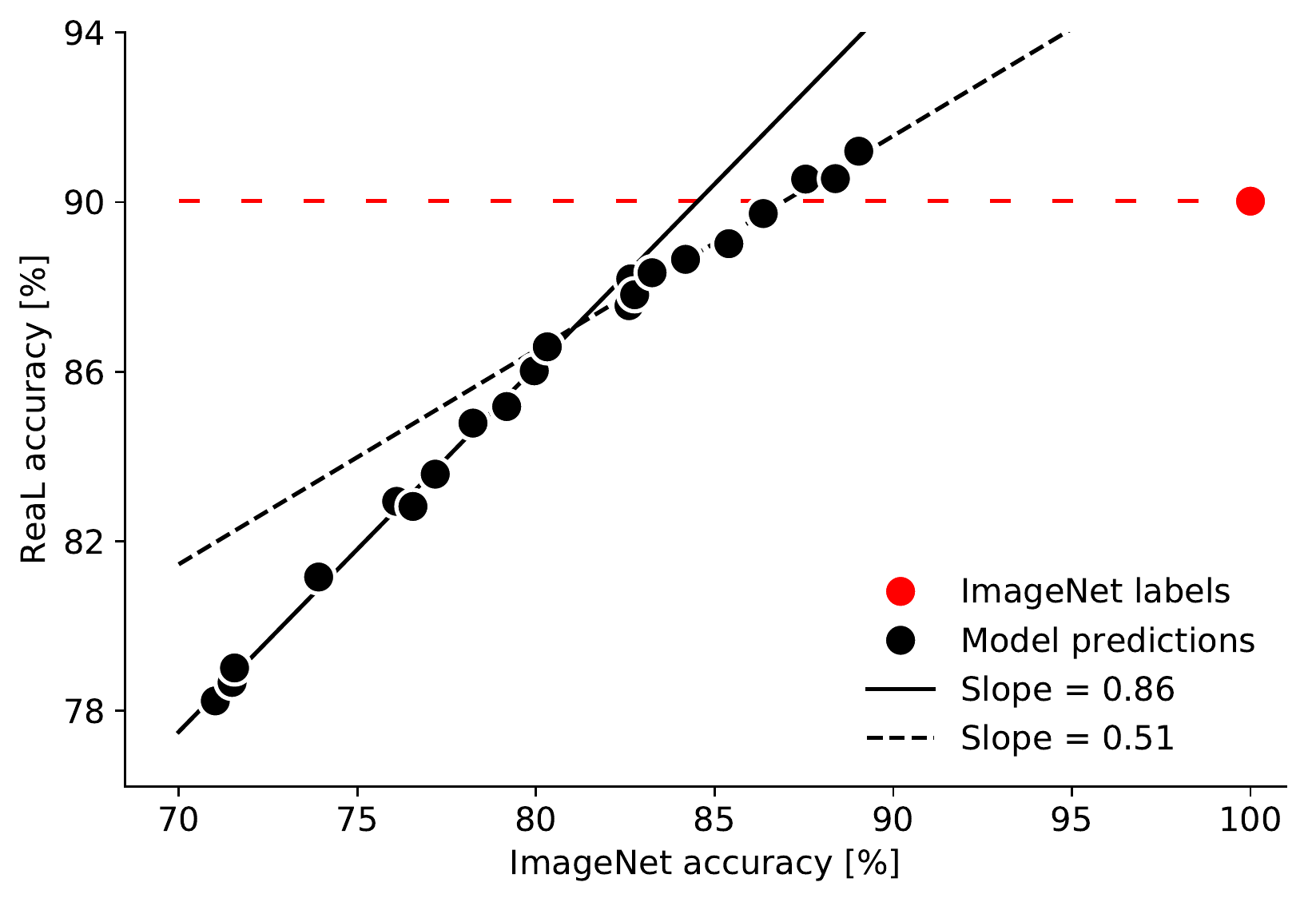}
  \end{center}
  \caption{Comparing progress on ReaL accuracy and the original ImageNet accuracy. We measured the association between both metrics by regressing ImageNet accuracy onto ReaL accuracy for the first (solid line) and second half (dashed line) of the models in our pool.}
  \vspace{-1em}
  \label{fig:association}
\end{wrapfigure}

Using the proposed ReaL accuracy, we now re-assess the progress of recently proposed classifiers on the ImageNet dataset.
In particular, given the strong generalization shown by early models (Section~\ref{sec:intro}) we would expect ImageNet accuracy to be very predictive of ReaL accuracy for these models.
This was indeed the case: when regressing ImageNet accuracy onto ReaL accuracy for the first half (ordered by ImageNet accuracy) of our models, we found a strong linear relationship (Figure~\ref{fig:association}, solid line: slope = 0.86).
When fitting the accuracies of more recent models (the second half), we again found a strong linear relationship, however its slope was significantly reduced (Figure~\ref{fig:association}, dashed line: slope = 0.51; $p$ < 0.001, Z-test on the difference in slopes). Importantly, several models \cite{kolesnikov2019big, xie2019self} already surpass the ReaL accuracy obtained by the original ImageNet labels (Figure~\ref{fig:association}, red point and dashed line). Together with the weakening relationship between ImageNet and ReaL accuracy, this suggests that we may be approaching the end of their utility as an evaluation metric.

In order to confirm this trend, we directly ask whether ReaL labels are better predicted by model outputs or ImageNet labels.
We search for all images in which the ImageNet label disagrees with a given model's prediction (i.e. ``mistakes'' according to the original ImageNet metric), and ask which is deemed correct by the ReaL labels. If the ReaL label considers both (model and ImageNet) predictions correct or incorrect, we discard the image.
We then compute the proportion of remaining images whose model-predicted label is considered correct.
The results of this analysis are very consistent with our previous finding: while early models are considerably worse than ImageNet labels at predicting human preferences, recent models now surpass them (Figure \ref{fig:teaser}). 

To see if these trends continue even further with higher ImageNet accuracy, we also created an ensemble of the three best models by combining their logits. The resulting model reaches 89.0\% in original ImageNet accuracy, and 91.20\% ReaL accuracy, furthering outperforming ImageNet labels according to human preferences (Figure \ref{fig:teaser}).

\subsection{Beyond single label predictions}

\begin{wrapfigure}{r}{0.4\textwidth}\vspace{-1.4em}
  \begin{center}
\vspace{-3em}
    \includegraphics[width=0.39\textwidth]{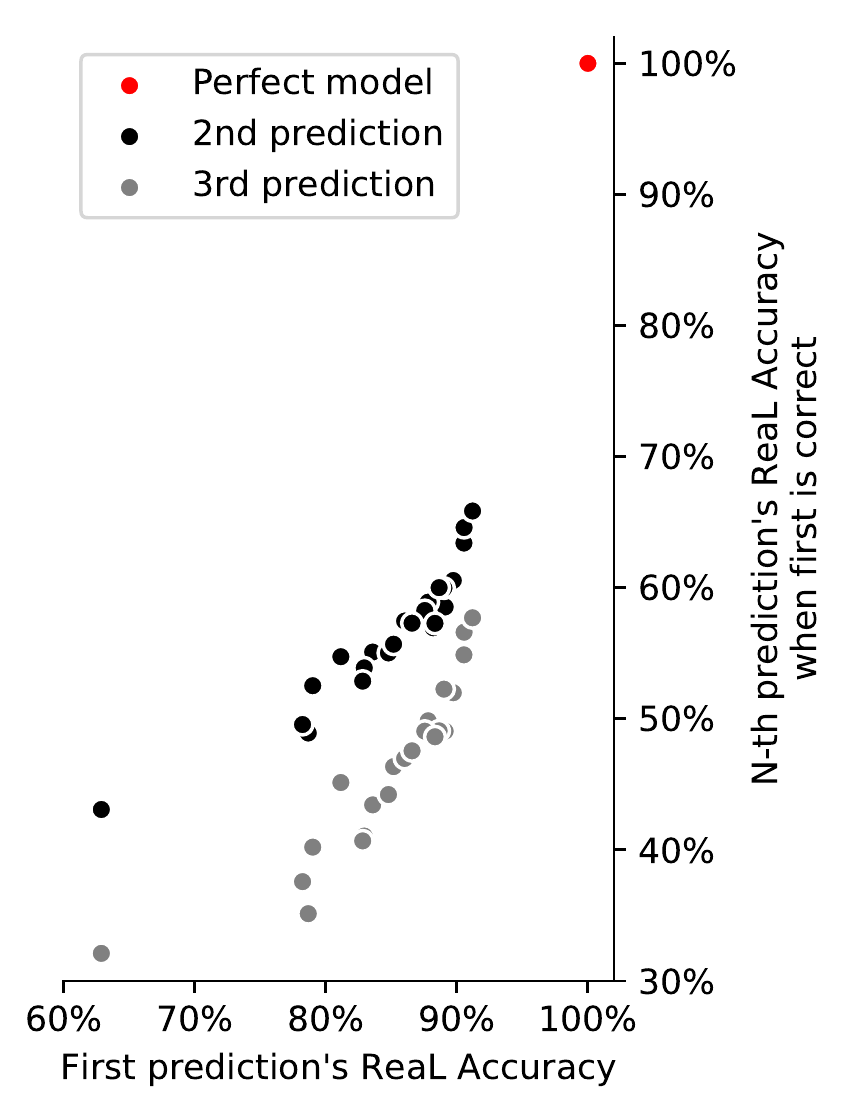}
  \end{center}\vspace{-1em}
  \caption{ReaL accuracy of models' \nth{2} and \nth{3} predictions.}
  \label{fig:top23}
  \vspace{-2em}
\end{wrapfigure}
The ReaL accuracy metric assesses whether a model's single most confident prediction falls within the set of acceptable answers. However, a more stringent criterion would ask whether all of the model's top predictions fall within this acceptable set. Given that our models are trained with a single label per image, these secondary predictions could very well be unconstrained and fairly meaningless. On the other hand, the visual similarity between classes could mean that an increase in accuracy of the top prediction will also entail an increase for later predictions.

Figure~\ref{fig:top23} shows that the accuracy of models' second and third predictions are significantly lower than their first.
Nevertheless, these predictions remain far superior to what would be expected due to chance, indicating that the correlations across classes enable meaningful secondary predictions.
Furthermore, these secondary accuracies display a striking correlation with the primary accuracy, indicating that it may be sufficient to improve upon the top-1 ReaL accuracy metric in order to also make progress on this more stringent task.
We therefore consider ReaL accuracy to be a good default metric to be used for our new set of labels.
Nevertheless, in the event that models start achieving near-perfect accuracy on this metric, it can easily be made more stringent by also accounting for secondary predictions.

\subsection{An analysis of co-occuring classes}

\begin{table}[t]
  \setlength{\tabcolsep}{0pt}
  \setlength{\extrarowheight}{5pt}
  \renewcommand{\arraystretch}{0.75}
  \centering
  \caption{ImageNet classes that often co-occur with other classes. The table presents class-level accuracies for different models. Current top-performing models outperform "Ideal" model that picks a correct label for each image at random. The last column shows top co-occurring classes, the number in brackets indicates percentage of how often a certain label co-occurs.}\label{tbl:cooc}
  \begin{tabularx}{\linewidth}{p{2.3cm}p{0.2cm}Cp{0.1cm}Cp{0.1cm}Cp{0.1cm}Cp{0.2cm}p{5.3cm}}
    \cmidrule[1pt]{1-11}
    \bf{ImageNet class} && \bf\begin{tabular}[c]{@{}c@{}}``Oracle''\\ model\end{tabular} && \bf\begin{tabular}[c]{@{}c@{}}Noisy\\ Student\end{tabular} && \bf\begin{tabular}[c]{@{}c@{}}Assemble\\ R152\end{tabular} && \bf{VGG16} && \bf{Top co-occuring classes} \\
    \cmidrule[0.5pt]{1-11}
    desktop comp. && 29.9\% && \bf\hspace{0.5em}71.1\%  && 71.1\% && 60.0\% && monitor (87\%) ; keyboard (78\%) \\
    muzzle        && 73.8\% && \bf100.0\% && 90.0\% && 55.0\% && german shepherd (5\%) ; holster (5\%) \\
    convertible   && 73.0\% && \bf\hspace{0.5em}97.6\%  && 95.2\% && 78.6\% && car wheel (40\%) ; grille (17\%) \\
    cucumber      && 74.2\% && \bf\hspace{0.5em}93.2\%  && 88.6\% && 70.5\% && zucchini (20\%) ; bell pepper (9\%) \\
    swing         && 87.5\% && \bf100.0\% && 94.0\% && 72.0\% && chain (12\%) ; sweatshirt (2\%) \\
    \cmidrule[1pt]{1-11}
  \end{tabularx}
  \vspace{-2em}
\end{table}

Among ImageNet images that we re-annotate with ReaL labels, approximately 29\% either contain multiple objects or a category that corresponds to multiple synonym labels in ImageNet (Figure~\ref{fig:example-images}). This raises a natural question of how ImageNet models perform on such images. Do they predict one of the correct labels at random, or do they learn to exploit biases in the labeling procedure to guess the ImageNet label? In this section, we use our ReaL labels to study this question in depth.

We start by considering all images whose ImageNet label is included in the ReaL labels.
Based on the ReaL labels, we compute the expected per-class accuracy of an ``unbiased oracle'' model, which predicts one of the ReaL labels uniformly at random.
If there is no bias in the original ImageNet labeling procedure, the per-class accuracies of this model represent the highest achievable accuracy. 

To focus on ambiguous classes which frequently co-occur with one another, we only consider classes for which the unbiased oracle achieves less than 90\% accuracy.
Some of these ``ambiguous'' classes could in fact be due to labeling noise, hence we also exclude fine-grained animal classes, as human raters frequently make mistakes on these.
We are left with 253 classes that fit these criteria.
These include ambiguous pairs such as (sunglass, sunglasses), (bathtub, tub), (promontory, cliff) and (laptop, notebook) (see Figure~\ref{fig:example-images}, bottom row), as well as classes which frequently appear together, e.g. (keyboard, desk), (cucumber, zucchini) and (hammer, nail), see (Figure~\ref{fig:example-images}, top row).

\begin{wrapfigure}{r}{0.6\textwidth}
  \begin{center}
  \vspace{-1.65em}
    \includegraphics[width=0.6\textwidth]{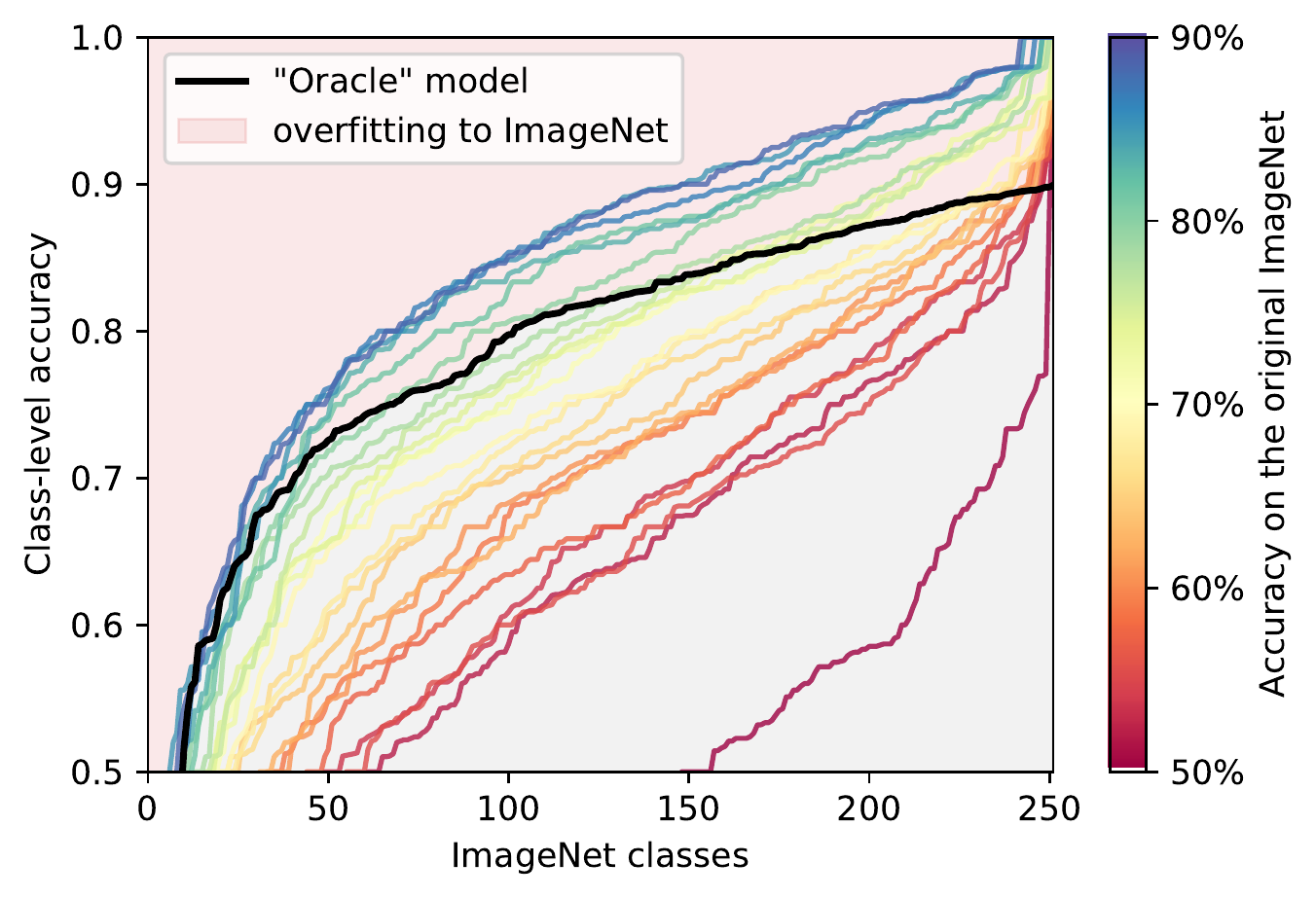}
  \end{center}
  \vspace{-0.75em}
  \caption{Each color curve corresponds to an ImageNet model and depicts sorted class-level accuracies. 
  The black curve depicts accuracies of the "unbiased oracle".
  Recent top-performing models dominate the oracle curve and, thus, are overfitting to label biases present in ImageNet.
  }
  \vspace{-2em}
  \label{fig:cooc}
\end{wrapfigure}

In Table~\ref{tbl:cooc} we illustrate this in more detail.
For instance, the ``desktop computer'' category frequently co-occurs with many other categories: more than 75\% of images that have a ``desktop computer'' also have a ``monitor'', a ``computer keyboard'' and a ``desk''.
If there was no label bias, the highest achievable accuracy on the ``desktop computer'' category would be approximately 30\%.
However, we observe that both the Noisy Student and the Assemble-ResNet152 models achieve a significantly higher accuracy of 71.1\%.

In Figure~\ref{fig:cooc} we compare the distribution of class-level accuracies of all models from our study to the unbiased oracle. The figure demonstrates that recent top-performing models, such as BiT-L and Noisy Student dominate the unbiased oracle model. This implies that a significant fraction of progress on ImageNet has been achieved by exploiting biases present in the ImageNet labeling procedure, and may explain why these gains only partially transfer to ReaL accuracy.

\section{Analyzing the remaining mistakes}

Even the highest-performing models display an error rate of approximately 11\% according to ImageNet labels, and 9\% according to ReaL labels. What is the nature of these mistakes, and what does this imply regarding the progress to be made on each of these benchmarks? 

\begin{wrapfigure}{r}{0.45\textwidth}
  \begin{center}
  \vspace{-2em}
    \includegraphics[width=0.44\textwidth]{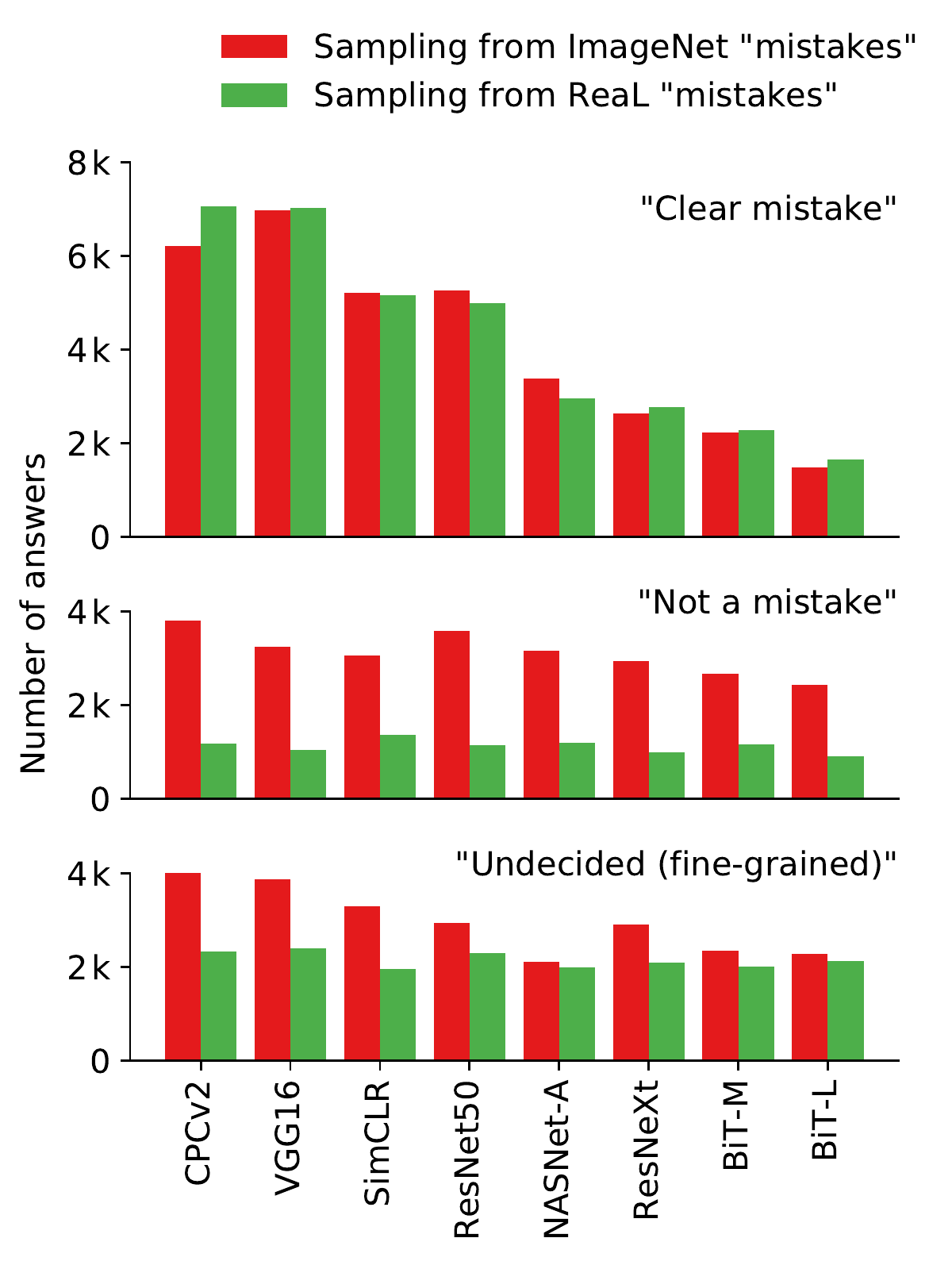}
  \end{center}
  \vspace{-1em}
  \caption{Remaining mistake types. We presented human raters with model predictions deemed incorrect according to either the ImageNet labels, or the ReaL labels, and asked them whether these predictions were indeed ``clear mistakes'', in fact ``not a mistake'', or simply ``undecidable''.}\label{fig:mistake_types}
  \vspace{-2em}
\end{wrapfigure}

To answer these questions, we design a follow-up study which shows images and model predictions that are ``mistakes'' according to a particular metric (i.e. ImageNet labels or ReaL labels).
To put these ``mistakes'' into context, we also show raters what the ``correct'' labels are for this image, as well as example images from those classes.
We then instruct raters to identify the reason for the prediction being considered a mistake: is it indeed a mistake?
Or is the prediction actually correct, and it is the label which is incorrect or incomplete?
Or is it neither of these options, for example if the labels are of fine-grained classes which the rater cannot distinguish?
\footnote{We had more options such as ``illegible image'' or ``depiction of A as B'', but found that they were not used.}

We sort the models used in this study according to their ImageNet accuracy. Using this ordering, we can appreciate a systematic decrease in the number of ``clear mistakes'', indicating that that there is indeed genuine progress on both benchmarks (Figure~\ref{fig:mistake_types}, top row). Nevertheless, there remains a significant fraction of ``mistakes'' which are deemed correct by human raters (Figure~\ref{fig:mistake_types}, middle row). Importantly, these cases are much more frequent when sampling from ImageNet mistakes than ReaL mistakes, indicating that the ReaL labels have indeed remedied a large portion of these false negatives.

Finally, a non-negligible number of mistakes are considered to be ``undecidable'' by raters (Figure~\ref{fig:mistake_types}, bottom row). Indeed, certain fine-grained classes are often hard or impossible to differentiate by non-experts. For these, ImageNet labels and ReaL labels yield similar assessments of high-performing models, indicating that collecting annotations from human experts may be necessary to measure progress along this axis.
In summary, while ReaL labels seem to be of comparable quality to ImageNet ones for fine-grained classes, they have significantly reduced the noise in the rest, enabling further meaningful progress on this benchmark.

\section{Improving ImageNet training}\label{sec:imp-imagenet-main}
\begin{table}[t]
  \setlength{\tabcolsep}{0pt}
  \setlength{\extrarowheight}{5pt}
  \renewcommand{\arraystretch}{0.75}
  \centering
  \caption{Top-1 accuracy (in percentage) on ImageNet with our proposed sigmoid loss and clean label set.
Median accuracy from three runs is reported for all the methods.
Either sigmoid loss or clean label set leads to consistent improvements over baseline. Using both achieves the best performance.
The improvement of our proposed method is more pronounced with longer training schedules.}\label{tbl:improved_training}
  \begin{tabularx}{\linewidth}{p{0.6cm}p{0.2cm}p{1.5cm}p{0.2cm}Cp{0.1cm}Cp{0.1cm}Cp{0.5cm}Cp{0.1cm}Cp{0.1cm}C}
    \cmidrule[1pt]{1-15}
    && \multirow{3}{=}[3pt]{\bf{Model}} && \multicolumn{5}{c}{\bf{ImageNet accuracy}} && \multicolumn{5}{c}{\bf{ReaL accuracy}} \\
    \cmidrule[0.5pt]{5-9} \cmidrule[0.5pt]{11-15}
    && && 90 epochs && 270 epochs && 900 epochs && 90 epochs && 270 epochs && 900 epochs \\

    \cmidrule[1pt]{1-15}
    \multirow{4}{=}{\centering \rotatebox{90}{ResNet-50}} && Baseline 
                 && 76.0 \phantom{\showdiff{(+9.9)}} && 76.9 \showdiff{(+0.9)} && 75.9 \showdiffn{(-0.1)}
                 && 82.5 \phantom{\showdiff{(+9.9)}} && 82.9 \showdiff{(+0.4)} && 81.6 \showdiffn{(-0.9)} \\
    && + Sigmoid && 76.3 \showdiff{(+0.3)} && 77.8 \showdiff{(+1.8)} && 76.9 \showdiff{(+0.9)}
                 && 83.0 \showdiff{(+0.5)} && 83.9 \showdiff{(+1.4)} && 82.7 \showdiff{(+0.2)} \\ 
    && + Clean   && 76.4 \showdiff{(+0.4)} && 77.8 \showdiff{(+1.8)} && 77.4 \showdiff{(+1.4)}
                 && 82.8 \showdiff{(+0.3)} && 83.7 \showdiff{(+1.2)} && 83.3 \showdiff{(+0.8)} \\ 
    && + Both    && 76.6 \showdiff{(+0.6)} && 78.2 \showdiff{(+2.2)} && \bf78.5 \showdiff{(+2.5)}
                 && 83.1 \showdiff{(+0.6)} && \bf84.3 \showdiff{(+1.8)} && 84.1 \showdiff{(+1.6)} \\

    \cmidrule[0.5pt]{1-15}
    \multirow{4}{=}{\centering \rotatebox{90}{ResNet-152}} && Baseline
                 && 78.0 \phantom{\showdiff{(+9.9)}} && 78.3 \showdiff{(+0.3)} && 77.1 \showdiffn{(-0.9)}
                 && 84.1 \phantom{\showdiff{(+9.9)}} && 83.8 \showdiffn{(-0.3)} && 82.3 \showdiffn{(-1.8)} \\
    && + Sigmoid && 78.5 \showdiff{(+0.5)} && 78.7 \showdiff{(+0.7)} && 77.4 \showdiffn{(-0.6)}
                 && 84.6 \showdiff{(+0.5)} && 84.3 \showdiff{(+0.2)} && 82.7 \showdiffn{(-1.4)} \\
    && + Clean   && 78.6 \showdiff{(+0.6)} && 79.6 \showdiff{(+1.6)} && 79.0 \showdiff{(+1.0)}
                 && 84.4 \showdiff{(+0.3)} && 85.0 \showdiff{(+0.9)} && 84.4 \showdiff{(+0.3)} \\
    && + Both    && 78.7 \showdiff{(+0.7)} && \bf79.8 \showdiff{(+1.8)} && 79.3 \showdiff{(+1.3)}
                 && 84.6 \showdiff{(+0.5)} && \bf85.2 \showdiff{(+1.1)} && 84.5 \showdiff{(+0.4)} \\

    \cmidrule[1pt]{1-15}
  \end{tabularx}
\vspace{-4mm}
\end{table}

We have identified two shortcomings of the ImageNet labeling procedure, both of which stem from assigning a single label to images which require multiple.
As ImageNet training images were annotated using the same process, we would expect similar types of noise in their labels as well.
In this section we investigate two ways of addressing this problem.

First, we propose using a training objective which allows models to emit multiple non-exclusive predictions for a given image.
For this we investigate treating the multi-way classification problem as a set of independent binary classification problems, and penalizing each one with a sigmoid cross-entropy loss, which does not enforce mutually exclusive predictions.

Second, based on our observation that recent top-performing models surpass the original ImageNet labels in their ability to predict human preferences (Figure~\ref{fig:teaser}), we asked whether we could use these models to filter the noise in ImageNet labels. 
We use one of the best models from our study, BiT-L, to clean the ImageNet training set.
In order to counter memorization of training data, we split training images into 10 equally-sized folds.
For each fold we perform the following: (1) Hold the fold out and train the BiT-L model on the remaining 9 folds and (2) Use the resulting model to predict labels on the hold-out fold.
We then remove all images whose labels are inconsistent with BiT-L's predictions.
Through this procedure, we retain approximately $90\%$ of the original training set. 

Finally, we investigate these two techniques in the context of longer training schedules. This is inspired by~\cite{arpit2017closer}, which observes that long training schedules can be harmful in the presence of noisy data.
We therefore expect cleaning the ImageNet training set (or using the sigmoid loss) to yield additional benefits in this regime.

In Table~\ref{tbl:improved_training}, we present comprehensive empirical evaluation of ResNet models trained on ImageNet, which includes investigation into loss type (softmax vs. sigmoid), longer training and cleaned training data. 
Inspecting the table we draw multiple insights:
(1) Training on clean ImageNet data consistently improves accuracy of the resulting model.
The gains are more pronounced when the standard softmax loss is used.
(2) Changing the softmax loss to the sigmoid loss also results in consistent accuracy improvements across all ResNet architectures and training settings.
(3) Combining clean data and sigmoid loss leads to further improvements.
(4) Finally, as opposed to the standard setting (softmax loss, full ImageNet data) which overfits for long schedules, training for longer using clean labels and sigmoid loss results in large gains.

A ResNet-50 architecture trained for 900 epochs with our proposed sigmoid loss on the clean training set achieves $78.5\%$ top-1 accuracy, which improves by $2.5\%$ over the standard ResNet-50 baseline ($76.0\%$) as well as the 900-epoch baseline.
Importantly, the gain in ReaL accuracy is $1.6\%$, despite the 900-epoch baseline heavily overfitting in terms of ReaL accuracy ($-0.9\%$)
We provide an extended table of results and plots in the Appendix~\ref{sec:imp-imagenet}.

\section{Conclusion}

In this work, we investigated whether recent progress on the ImageNet benchmark amounts to meaningful generalization, or whether we have begun to overfit to artifacts in the validation set. To that end we identified a set of deficiencies in the original validation labels, and collected a new set of annotations which address these. Using these ``Reassessed Labels'' (ReaL), we found the association between progress on ImageNet and ReaL progress to have weakened over time, although it remains significant. Importantly, recent models have begun to surpass the original ImageNet labels in terms of their ReaL accuracy, indicating that we may be approaching the end of their usefulness. In a follow-up experiment, we found ReaL labels to remove more than half of the ImageNet labeling mistakes, implying that ReaL labels provide a superior estimate of the model accuracy.

Given the shortcomings of ImageNet labels we identified, we proposed two modifications to the canonical supervised training setup aimed at addressing them. Together, these simple modifications yield relatively large empirical gains, and indicate that label noise could have been a limiting factor for longer training schedules. 
Together, these findings suggest that although the original set of labels may be nearing the end of their useful life, ImageNet and its ReaL labels can readily benchmark progress in visual recognition for the foreseeable future.

\textbf{Acknowledgements.} We would like to thank every one of the raters involved in this task for their hard work, as well as their manager for making this a smooth experience. We also thank Pierre Ruyssen for engineering support in the initial phase of the project, and Ilya Tolstikhin for references and inspiration regarding the longer training schedules in Section~\ref{sec:imp-imagenet-main}. Finally, we are thankful to Tobias Weyand for his thorough review of a draft of this paper.

\bibliographystyle{unsrt}
\bibliography{neurips}

\newpage

\input{appendix}

\end{document}

%% file: appendix.tex
\appendix

\section{Details on proposal generation}
\label{sec:proposals}

The exact algorithm for the label proposal rule is as follows:
for each model, compute its logits and probabilities for each image, resulting in two times 50\,M image-label pairs.
From these, keep those that fall into the largest 150\,k logits or the largest 150\,k probabilities.
Pool all of these from all models.
Remove image-label pairs that have only been brought up a single time by a single model.
For each image, unconditionally add each model's top-1 prediction for each image, as well as the original ILSVRC-2012 label.

As mentioned in the main text, we first generate an exhaustive list of image-label proposals using all models shown in Table~\ref{tbl:models}, have 5 computer vision experts label a subset of images using these proposals, and then select a subset of models (indicated in Table~\ref{tbl:models}, ``Final'' column) that reaches an almost-perfect recall (97.08\%) and significantly increases precision (from 14.55\% to 28.33\%) in order to reduce the labeling burden.

\begin{table}[h]
  \setlength{\tabcolsep}{0pt}
  \renewcommand{\arraystretch}{1.25}
  \rowcolors{2}{gray!8}{white}
  \centering
  \caption{The models used for label proposal in this study, and their characteristics.}\label{tbl:models}
  \begin{tabularx}{\linewidth}{p{0.7cm}p{0.2cm}p{4cm}p{0.2cm}Cp{0.2cm}Cp{0.2cm}Cp{0.2cm}C}
    \cmidrule[1pt]{1-11}
    \bf{Ref.} && \bf{Model} && \bf{Arch. search} && \bf{Self-sup.} && \bf{Ext. data} && \bf{Final} \\

    \cmidrule[1pt]{1-11}
    \cite{simonyan2014very} && VGG-16 && no && no && no && yes \\
    \cite{szegedy2016rethinking} && Inception v3 && no && no && no && yes \\
    \cite{he2016deep} && ResNet-50 && no && no && no && no \\
    \cite{he2016deep} && ResNet-152 && no && no && no && no \\
    \cite{xie2017aggregated} && ResNeXt-101, 32x8d && no && no && no && no \\
    \cite{mahajan2018exploring} && ResNeXt-101, 32x8d, IG && no && no && yes && yes \\
    \cite{mahajan2018exploring} && ResNeXt-101, 32x48d, IG && no && no && yes && no \\
    \cite{kolesnikov2019big} && BiT-M && no && no && yes && yes \\
    \cite{kolesnikov2019big} && BiT-L && no && no && yes && yes \\
    \cite{lee2020compounding} && Assemble ResNet-50 && yes && no && no && no \\
    \cite{lee2020compounding} && Assemble ResNet-152 && yes && no && no && no \\
    \cite{zoph2018learning} && NASNet-A Large && yes && no && no && no \\
    \cite{zoph2018learning} && NASNet-A Mobile && yes && no && no && no \\
    \cite{cai2019once} && Once for all (Large) && yes && no && no && no \\
    \cite{zhai2019s4l} && S4L MOAM && no && yes && no && no \\
    \cite{henaff2019data} && CPC v2, fine-tuned && no && yes && no && yes \\
    \cite{henaff2019data} && CPC v2, linear && no && yes && no && no \\
    \cite{chen2020improved} && MoCo v2, long && no && yes && no && no \\
    \cite{chen2020simple} && SimCLR && no && yes && no && no \\
    \cmidrule[1pt]{1-11}
  \end{tabularx}
\end{table}



\clearpage

\section{Improving ImageNet training}\label{sec:imp-imagenet}

\begin{table}[h]
  \setlength{\tabcolsep}{0pt}
  \setlength{\extrarowheight}{5pt}
  \renewcommand{\arraystretch}{0.75}
  \centering
  \caption{Top-1 accuracy (in percentage) on ImageNet with our proposed sigmoid loss and clean label set.
Median accuracy from three runs is reported for all the methods.
Either sigmoid loss or clean label set leads to consistent improvements over baseline. Using both achieves the best performance.
The improvement of our proposed method is more pronounced with longer training schedules.}\label{tbl:improved_training2}
  \begin{tabularx}{\linewidth}{p{0.6cm}p{0.2cm}p{1.5cm}p{0.2cm}Cp{0.1cm}Cp{0.1cm}Cp{0.5cm}Cp{0.1cm}Cp{0.1cm}C}
    \cmidrule[1pt]{1-15}
    && \multirow{3}{=}[3pt]{\bf{Model}} && \multicolumn{5}{c}{\bf{ImageNet accuracy}} && \multicolumn{5}{c}{\bf{ReaL accuracy}} \\
    \cmidrule[0.5pt]{5-9} \cmidrule[0.5pt]{11-15}
    && && 90 epochs && 270 epochs && 900 epochs && 90 epochs && 270 epochs && 900 epochs \\

    \cmidrule[1pt]{1-15}
    \multirow{4}{=}{\centering \rotatebox{90}{ResNet-50}} && Baseline 
                 && 76.0 \phantom{\showdiff{(+9.9)}} && 76.9 \showdiff{(+0.9)} && 75.9 \showdiffn{(-0.1)}
                 && 82.5 \phantom{\showdiff{(+9.9)}} && 82.9 \showdiff{(+0.4)} && 81.6 \showdiffn{(-0.9)} \\
    && + Sigmoid && 76.3 \showdiff{(+0.3)} && 77.8 \showdiff{(+1.8)} && 76.9 \showdiff{(+0.9)}
                 && 83.0 \showdiff{(+0.5)} && 83.9 \showdiff{(+1.4)} && 82.7 \showdiff{(+0.2)} \\ 
    && + Clean   && 76.4 \showdiff{(+0.4)} && 77.8 \showdiff{(+1.8)} && 77.4 \showdiff{(+1.4)}
                 && 82.8 \showdiff{(+0.3)} && 83.7 \showdiff{(+1.2)} && 83.3 \showdiff{(+0.8)} \\ 
    && + Both    && 76.6 \showdiff{(+0.6)} && 78.2 \showdiff{(+2.2)} && \bf78.5 \showdiff{(+2.5)}
                 && 83.1 \showdiff{(+0.6)} && \bf84.3 \showdiff{(+1.8)} && 84.1 \showdiff{(+1.6)} \\

    \cmidrule[0.5pt]{1-15}
    \multirow{4}{=}{\centering \rotatebox{90}{ResNet-101}} && Baseline
                 && 77.4 \phantom{\showdiff{(+9.9)}} && 77.6 \showdiff{(+0.2)} && 76.4\showdiffn{(-1.0)}
                 && 83.7 \phantom{\showdiff{(+9.9)}} && 83.6 \showdiffn{(-0.1)} && 81.9 \showdiffn{(-1.8)} \\
    && + Sigmoid && 77.8 \showdiff{(+0.4)} && 78.2 \showdiff{(+0.8)} && 77.3 \showdiffn{(-0.1)}
                 && 84.2 \showdiff{(+0.5)} && 84.2 \showdiff{(+0.5)} && 82.7 \showdiffn{(-1.0)} \\
    && + Clean   && 77.9 \showdiff{(+0.5)} && 79.2 \showdiff{(+1.8)} && 78.6 \showdiff{(+1.2)}
                 && 84.0 \showdiff{(+0.3)} && 84.6 \showdiff{(+0.9)} && 84.0 \showdiff{(+0.3)} \\
    && + Both    && 78.0 \showdiff{(+0.6)} && \bf79.5 \showdiff{(+2.1)} && 79.2 \showdiff{(+1.8)}
                 && 84.3 \showdiff{(+0.6)} && \bf85.1 \showdiff{(+1.4)} && 84.4 \showdiff{(+0.7)} \\
                 
    \cmidrule[0.5pt]{1-15}
    \multirow{4}{=}{\centering \rotatebox{90}{ResNet-152}} && Baseline
                 && 78.0 \phantom{\showdiff{(+9.9)}} && 78.3 \showdiff{(+0.3)} && 77.1 \showdiffn{(-0.9)}
                 && 84.1 \phantom{\showdiff{(+9.9)}} && 83.8 \showdiffn{(-0.3)} && 82.3 \showdiffn{(-1.8)} \\
    && + Sigmoid && 78.5 \showdiff{(+0.5)} && 78.7 \showdiff{(+0.7)} && 77.4 \showdiffn{(-0.6)}
                 && 84.6 \showdiff{(+0.5)} && 84.3 \showdiff{(+0.2)} && 82.7 \showdiffn{(-1.4)} \\
    && + Clean   && 78.6 \showdiff{(+0.6)} && 79.6 \showdiff{(+1.6)} && 79.0 \showdiff{(+1.0)}
                 && 84.4 \showdiff{(+0.3)} && 85.0 \showdiff{(+0.9)} && 84.4 \showdiff{(+0.3)} \\
    && + Both    && 78.7 \showdiff{(+0.7)} && \bf79.8 \showdiff{(+1.8)} && 79.3 \showdiff{(+1.3)}
                 && 84.6 \showdiff{(+0.5)} && \bf85.2 \showdiff{(+1.1)} && 84.5 \showdiff{(+0.4)} \\

    \cmidrule[1pt]{1-15}
  \end{tabularx}
\vspace{-4mm}
\end{table}

 \begin{figure}[h]
   \begin{center}
     \includegraphics[width=1.0\textwidth]{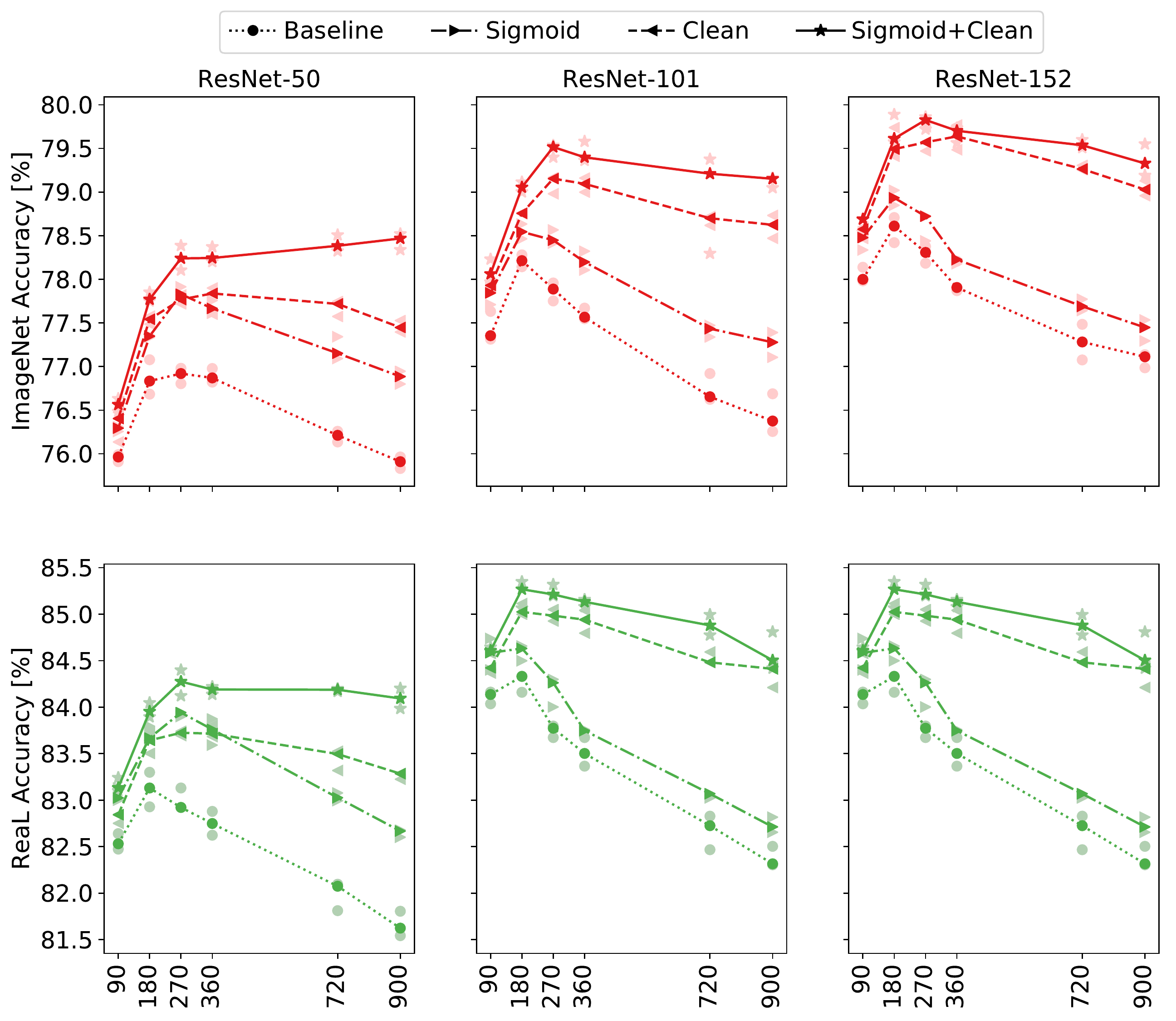}
   \end{center}
   \caption{Top-1 accuracy (in percentage) on ImageNet with our proposed sigmoid loss and clean label set. 
   X-axis shows the training epochs. 
   The above row shows results on the original ImageNet labels, and the bottom row shows the ReaL accuracy for the models. 
   The curve shows the median results across three runs, where we use light color to mark the other two runs.
   Either sigmoid loss or clean label set leads to consistent improvements over the baseline (standard cross-entropy loss with softmax on the all original ImageNet images) and
   using both results in the best accuracies.}
 \end{figure}

\clearpage

\section{ReaL accuracies of all models in Fig.~\ref{fig:association}}

\begin{table}[h]
  \setlength{\tabcolsep}{0pt}
  \renewcommand{\arraystretch}{1.25}
  \rowcolors{2}{gray!8}{white}
  \centering
  \captionsetup{width=0.8\linewidth}
  \caption{ReaL and ``original'' ILSVRC-2012 top-1 accuracies (in percent) of all models used for Fig.~\ref{fig:association} in the main paper, sorted by their ReaL accuracies.}\label{tbl:accuracies}
  \begin{tabularx}{0.8\linewidth}{p{0.7cm}p{0.2cm}p{5cm}p{0.2cm}Cp{0.2cm}C}
    \cmidrule[1pt]{1-7}
    \bf{Ref.} && \bf{Model} && \bf{ReaL Acc.} && \bf{Orig. Acc.} \\

    \cmidrule[1pt]{1-7}
    self                           && Top-3 Ensemble \cite{xie2019self,kolesnikov2019big,touvron2019fixing} && 91.20 && 89.03 \\
    \cite{xie2019self}             && NoisyStudent-L2             && 90.55 && 88.38 \\
    \cite{kolesnikov2019big}       && BiT-L                       && 90.54 && 87.55 \\
    \cite{russakovsky2015imagenet} && ILSVRC-2012 labels          && 90.02 && 100.0 \\
    \cite{touvron2019fixing}       && Fix-ResNeXt-101, 32x48d, IG && 89.73 && 86.36 \\
    \cite{mahajan2018exploring}    && ResNeXt-101, 32x48d, IG     && 89.11 && 85.40 \\
    \cite{kolesnikov2019big}       && BiT-M                       && 89.02 && 85.40 \\
    \cite{lee2020compounding}      && Assemble ResNet-152         && 88.65 && 84.19 \\
    \cite{henaff2019data}          && CPC v2, fine-tuned (100\%)  && 88.33 && 83.25 \\
    \cite{mahajan2018exploring}    && ResNeXt-101, 32x8d, IG      && 88.19 && 82.65 \\
    \cite{lee2020compounding}      && Assemble ResNet-50          && 87.82 && 82.76 \\
    \cite{zoph2018learning}        && NASNet-A Large              && 87.56 && 82.60 \\
    \cite{zhai2019s4l}             && S4L MOAM                    && 86.59 && 80.32 \\
    \cite{cai2019once}             && Once for all (Large)        && 86.02 && 79.95 \\
    \cite{xie2017aggregated}       && ResNeXt-101, 32x8d          && 85.18 && 79.18 \\
    \cite{he2016deep}              && ResNet-152                  && 84.79 && 78.24 \\
    \cite{szegedy2016rethinking}   && Inception v3                && 83.58 && 77.18 \\
    \cite{zhai2019s4l}             && S4L-Rotation (100\%)        && 83.10 && 76.72 \\
    \cite{he2016deep}              && ResNet-50                   && 82.94 && 76.10 \\
    \cite{chen2020simple}          && SimCLR                      && 82.82 && 76.56 \\
    \cite{zhai2019s4l}             && S4L-Exemplar (100\%)        && 81.26 && 74.41 \\
    \cite{zoph2018learning}        && NASNet-A Mobile             && 81.15 && 73.92 \\
    n/a                            && VGG-16 + BatchNorm          && 80.60 && 73.34 \\
    \cite{zhai2019s4l}             && S4L MOAM (10\%)             && 80.41 && 73.22 \\
    \cite{henaff2019data}          && CPC v2, fine-tuned (10\%)   && 80.28 && 72.90 \\
    \cite{simonyan2014very}        && VGG-16                      && 79.01 && 71.56 \\
    \cite{henaff2019data}          && CPC v2, linear              && 78.67 && 71.49 \\
    \cite{chen2020improved}        && MoCo v2, long               && 78.23 && 71.02 \\
    \cite{zhai2019s4l}             && S4L-Exemplar (10\%)         && 69.76 && 62.21 \\
    \cite{zhai2019s4l}             && S4L-Rotation (10\%)         && 69.05 && 61.37 \\
    \cite{krizhevsky2012imagenet}  && AlexNet                     && 62.88 && 56.36 \\
    \cmidrule[1pt]{1-7}
  \end{tabularx}
\end{table}